\documentclass{article}

 \usepackage[main, final]{neurips_2026}
\usepackage[utf8]{inputenc} % allow utf-8 input
\usepackage[T1]{fontenc}    % use 8-bit T1 fonts
\usepackage[hidelinks]{hyperref}       % hyperlinks
\usepackage{url}            % simple URL typesetting
\usepackage{booktabs}       % professional-quality tables
\usepackage{amsfonts}       % blackboard math symbols
\usepackage{nicefrac}       % compact symbols for 1/2, etc.
\usepackage{microtype}      % microtypography
\usepackage{xcolor}         % colors
\usepackage[]{todonotes}
\usepackage{amsmath}
\usepackage{amssymb}
\usepackage{amsthm}
\usepackage{nicefrac}
\usepackage{microtype}
\usepackage{xcolor}
\usepackage[ruled,vlined,linesnumbered]{algorithm2e}
\usepackage{wrapfig}        % half-width Algorithm 1 (camera-ready)
\usepackage{subcaption}
\usepackage{multirow}
\usepackage{enumitem}
\usepackage{cleveref}
\usepackage{pgfplots}
\pgfplotsset{compat=1.18}
\usepgfplotslibrary{groupplots}
\usepackage{tikz}
\usetikzlibrary{arrows.meta, positioning, shapes.geometric, backgrounds, fit, calc, decorations.pathmorphing}
\usepackage{ltl}

\newcommand{\rob}{\rho}
\newcommand{\trace}{\sigma}
\newcommand{\tbt}{\mathcal{T}}
\newcommand{\stlf}{\varphi}
\newcommand{\tbtcl}{\textsc{TBTCL}}

\newcommand{\nn}[1]{\textnormal{\textsc{#1}}}
\newtheorem{definition}{Definition}
\newtheorem{proposition}{Proposition}
\newtheorem{lemma}{Lemma}
\newtheorem{corollary}{Corollary}

\title{Temporal Behavior Trees for Reinforcement Learning: Specifications as Rewards, Curricula, and Diagnostics}

\newcommand{\rev}[1]{{\color{red}#1}}

\author{%
  Vladimir Krsmanovi\'{c}$^{1}$\thanks{Corresponding Author}
  \qquad
  Bernd Finkbeiner$^{1,2}$ \\
  $^{1}$\textit{CISPA Helmholtz Center for Information Security} \quad
  $^{2}$\textit{Technical University of Munich} \\
  \texttt{\{vladimir.krsmanovic,finkbeiner\}}@cispa.de
}

\begin{document}

\maketitle

\begin{abstract}
  Reinforcement learning from formal specifications often separates task definition, reward shaping, curriculum design, and policy diagnostics into different mathematical objects. This fragmentation requires additional conversion steps and discards structure already present in the specification, such as subtask boundaries, ordering constraints, and quantitative measures. We introduce TBTCL~(Temporal Behavior Tree Centric Learning), a framework that uses Temporal Behavior Trees (TBTs) as a unified interface for specification-guided reinforcement learning. A single TBT specifies the task, induces dense robustness-based rewards, defines a curriculum, and supports better understanding of learned policies. By adapting TBT syntax and semantics for specification-guided reinforcement learning, we provide a formalism that is more expressive than popular specification languages such as LTL and inherently hierarchical. We present algorithms to automatically extract a curriculum DAG and derive dense, robustness-based rewards directly from TBT semantics. Furthermore, we utilize TBT trace segmentation to build understandable competence profiles and automatically adapt the reward toward bottlenecks. We evaluate TBTCL on six specifications across discrete and continuous domains and show that Temporal Behavior Trees can serve as a specification, a reward, a curriculum, and a diagnostic.

\end{abstract}

\section{Introduction}
\label{sec:intro}
Temporal Logics are a powerful tool in reinforcement learning, enabling the formal specification of tasks and serving as the engine of Specification-Guided Reinforcement Learning. Most standard approaches convert a logical specification into a deterministic finite automaton, which is then used to guide the learning process. While this conversion is necessary to translate standard logics such as LTL~\citep{pnueli1977temporal} into a form suitable for RL, it incurs a high cost as it loses crucial information at each step. Properties such as subtask boundaries, quantitative performance measures, hierarchical structures, and task orderings and priorities are already present in the logical specification but are discarded by most methods during intermediate conversion steps.

The core cause of these problems is that most specification languages are not designed for purposes beyond task specification. A typical pipeline represents the task specification in one object, derives a reward in another, constructs a curriculum through an additional procedure and yet another representation, and then relies on separate diagnostics and derivations to understand how a learned policy behaves, performs, and fails. 
As such, a potential solution would be a specification language that provides the task definition, a dense reward, a curriculum, and behavioral diagnostics, without requiring separate conversion algorithms, representations, or user inputs for each. In this paper, we argue that a single formal object can simultaneously serve as the task definition, a dense reward source, a curriculum structure, and a diagnostic model.

We show that Temporal Behavior Trees (TBTs)~\citep{schirmer2024tbt} have these important properties and provide a flexible formalism for reinforcement learning. Existing work has developed this formalism for monitoring and trace analysis of cyber-physical systems~\citep{schirmer2024tbt,schirmer2026specifying,schirmer2026trace} and has demonstrated its power and flexibility in these applications.

A TBT composes STL formulas at its leaves with tree-level operators such as sequence, parallel composition, and fallback. The STL leaves provide quantitative robustness, while the tree structure describes how behaviors compose over a trace. This hierarchical structure, which combines STL formulas at the leaves with tree-level operators, enables a concise and expressive definition of subtasks and their interconnections, which are frequently goals in reinforcement learning. This combination further enables more concise and expressive specifications than those of commonly used logics such as LTL. Furthermore, the quantitative semantics encoded by a robustness metric for TBT yields a dense reward signal and builds on previous work in the field that adapted the robustness of logics such as STL for Reinforcement Learning~\citep{varnai2020stl}. Finally, TBTs support trace decomposition, connecting policy rollouts directly to the specification expressed as a TBT without additional steps, and can be used to evaluate policy behavior and identify bottlenecks and weak points. 

Building on these properties, we adapt the Temporal Behavior Trees for reinforcement learning and demonstrate their use in specification, training, and evaluation. We present \tbtcl{}~(Temporal Behavior Tree Centric Learning), a framework that exploits this unified structure. Given a TBT specification, \tbtcl{} extracts a curriculum represented by a directed acyclic graph~(DAG) by structural recursion over the tree, trains a single policy through a bottom-up sequence of subformula reward functions, and advances stages based on achieved STL robustness rather than fixed episode budgets. When learning stalls, trace segmentation can identify the bottleneck subtask and redirect training to address it. Moreover, the same segmentation algorithm that drives online curriculum adaptation can also serve as a runtime diagnostic mechanism, decomposing any evaluation trace into per-subtask quality scores directly aligned with TBT semantics, without requiring any additional external mechanism.

\begin{figure}[t]
  \centering
  \includegraphics[width=0.95\linewidth]{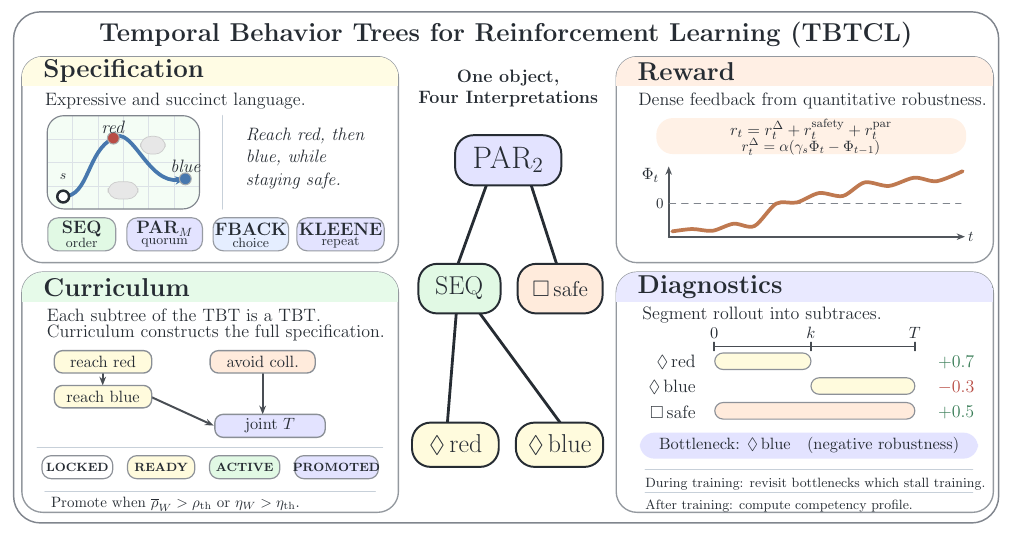}
  \caption{Temporal Behavior Trees provide a single formal object for task specification, dense robustness-based reward shaping, curriculum extraction, and trace-based diagnostics.}
  \vspace{-0.5cm}
  \label{fig:tbtcl-overview}
\end{figure}

We make the following contributions:
 \begin{itemize}[nosep,leftmargin=*]
      \item We present Temporal Behavior Trees as a unified interface for specification-guided RL, adapting the semantics to reinforcement learning (Section~\ref{sub:tbt-spec}).
      \item We use the quantitative robustness semantics of TBTs to derive a dense reward from the prefix robustness of the TBT (Section~\ref{sub:tbt-reward}).
      \item We leverage the tree structure of TBTs and present an algorithm to extract a curriculum directly from their structure and semantics (Section~\ref{sub:tbt-curr}).
      \item We demonstrate how the TBT trace segmentation mechanism can be used to understand policies and improve their learning by identifying bottlenecks during training (Section~\ref{sub:tbt-diag}).
\end{itemize}

\section{Background}
\label{sec:background}

\paragraph{Signal Temporal Logic.}
Signal Temporal Logic (STL)~\citep{maler2004stl,donze2010robust} was developed for monitoring and verification of cyber-physical systems. 

An STL formula is defined inductively as $\stlf \;{::=}\; \mu \;\mid\; \neg\stlf \;\mid\; \stlf_1 \wedge \stlf_2 \;\mid\; \stlf_1 \vee \stlf_2 \;\mid\; \Diamond_{[l,u]}\stlf \;\mid\; \Box_{[l,u]}\stlf \;\mid\; \stlf_1 \,\mathcal{U}_{[l,u]}\, \stlf_2$, where $\mu: \mathbb{R}^n \to \mathbb{R}$ is an atomic predicate over the state space and $[l,u]$ are optional time bounds (omitted $\Rightarrow$ full signal). $\Diamond$ (Eventually) requires satisfaction at some point; $\Box$ (Globally) requires it at all points; $\mathcal{U}$ (Until) requires the first operand to hold until the second becomes true.

It extends Boolean temporal logic to real-valued signals, replacing the binary satisfied/violated verdict with a \emph{quantitative robustness} score. The robustness of a given STL formula is a real number whose sign indicates satisfaction and whose magnitude measures the margin.  We provide a full definition of STL semantics, syntax and recursive robustness computation in Appendix~\ref{app:definitions}.

\paragraph{Temporal Behavior Trees.}
Temporal Behavior Trees (TBTs)~\citep{schirmer2024tbt} compose STL leaves with hierarchical operators, yielding a two-level language: tree structure for compositional organization, STL for quantitative leaf-level properties. A TBT leaf $\nn{Leaf}(\stlf)$ evaluates an STL formula, while a sequence node $\nn{Seq}(\tbt_1,\tbt_2)$, a parallel node $\nn{Par}_M(\tbt_1,\ldots,\tbt_k)$ with quorum $M$, and a fallback node $\nn{Fback}(\tbt_1,\ldots,\tbt_k)$ compose child behaviors over trace segments:
\[
\tbt ::= \nn{Leaf}(\stlf)
\mid \nn{Seq}(\tbt_1,\tbt_2)
\mid \nn{Par}_M(\tbt_1,\ldots,\tbt_k)
\mid \nn{Fback}(\tbt_1,\ldots,\tbt_k).
\]
Intuitively, $\nn{Seq}$ searches for a cut point such that the prefix satisfies the first child and the suffix satisfies the second. $\nn{Par}_M$ evaluates all children on the same segment and requires at least $M$ to succeed (quorum). $\nn{Fback}$ represents alternatives, selecting the child and suffix with maximal robustness. We provide a detailed description of the TBT syntax and semantics in Appendix~\ref{app:definitions}.

\paragraph{Rollouts as traces.}
We consider episodic reinforcement learning in a Markov Decision Process (MDP) with horizon $H$ and initial-state distribution $\mu_0$. A rollout under policy $\pi$, starting from $s_0\mathrel{\sim}\mu_0$, induces a finite trajectory $\trace = (s_0,\ldots,s_H)$ of environment states, which we call a \emph{trace} following the formal methods literature, optionally augmented with real-valued signals such as distances to goals, hazards, or objects. STL predicates are evaluated on this trace: an atomic predicate $\mu$ maps a state or signal value to a real number, with $\mu(s_t)>0$ indicating that the predicate is satisfied at time $t$. We write $\trace[l{:}u]$ to denote the contiguous segment from time $l$ to $u$.

\section{Method}
We show that  Temporal Behavior Trees can act as the specification \emph{and} as a dense reward \emph{and} as a curriculum \emph{and} as a diagnostic metric.
First, in Section~\ref{sub:tbt-spec}, we argue that TBTs are an expressive and succinct specification language for describing complex behaviors and a powerful formalism for describing reinforcement learning tasks. Afterward, in Section~\ref{sub:tbt-reward}, we show how STL's robustness metrics are inherited by TBTs and present adaptations that enable effective training directly using TBT semantics. In Section~\ref{sub:tbt-curr} we demonstrate how the tree structure of TBTs directly induces a curriculum schedule, and show theoretical results about it. Finally, in Section~\ref{sub:tbt-diag} we use TBTs' ability to segment traces (in our case, policy rollouts) to both better understand learned policies and guide the learning process towards better ones.

\subsection{Temporal Behavior Trees as Specifications}\label{sub:tbt-spec}
Temporal Behavior Trees provide an expressive and concise specification language, owing to their hierarchical structure and expressiveness of the underlying semantics. Temporal Behavior Trees provide not only a more succinct way to write out many specifications, but are also known to be more expressive than other commonly used logics. As shown by \citet{schirmer2024tbt} building on proofs from~\citet{wolper1983temporal}, Temporal Behavior Trees are strictly more expressive than STL, whose robustness semantics they inherit.

Each leaf of a TBT contains an STL formula describing a simple task, such as reaching a target or avoiding a hazard.
Internal nodes specify how behaviors described by child nodes compose: a Sequence node~(\nn{Seq}) imposes temporal order (\textit{reach A before reaching B}), a parallel node~($\nn{Par}_M$) defines simultaneous requirements (\textit{reach  A while avoiding C}), and a Fallback node~($\nn{FBack}$) represents alternatives (\textit{reach either target A or target B}). The combination of simple-to-specify leaf properties with powerful operators enables users to specify hierarchical specifications rather than just flat, linear-time formulas.

We further adapt the TBT formalism by~\citet{schirmer2024tbt} for reinforcement learning by providing an operator to describe repeated behaviors, $\nn{Kleene}(n,\tbt_c)$~. This node enables the specification of $n{+}1$ successful repetitions of the child behavior $\tbt_c$. Its robustness can be computed as the best partition of the trace into $n{+}1$ segments, scored by the weakest repetition:
\begin{equation}
\rob(\trace[l{:}u], \nn{Kleene}(n,\tbt_c))
 =
\max_{\mathcal{P}_{n+1}(l,u)}
\min_{i=1}^{n+1} \rob(\trace_i,\tbt_c),
\label{eq:rob-kleene}
\end{equation}
where $\mathcal{P}_{n+1}(l,u)$ is the set of valid ordered partitions of $\trace[l{:}u]$ into $n{+}1$ non-empty contiguous segments. This operator enables easier specification of recurring properties (\textit{such as reach A 5 times}) and avoids the need to define the same behavior through lengthy unfoldings of a flat temporal formula~\footnote{The original TBT formalization by \citet{schirmer2024tbt} includes a $\star_n$ node that accepts \textit{up to} $n$ repetitions of a given behavior. While these semantics are a natural fit for runtime monitoring, they are less so for reinforcement learning, as \textit{up to} specifications are trivial to satisfy - which is why we replace this node with $\nn{Kleene}$}.%This facilitates the easier formalization of relevant specifications and enables the curriculum and diagnostic algorithms we present in later sections.

As a concrete example of one task description, in Figure~\ref{fig:tbt-vs-ltl} we show an example specification in a Temporal Behavior Tree as well as its equivalent in a formula in Linear Time Temporal Logic~(LTL)~\citep{pnueli1977temporal}. Due to the linear nature of LTL, many such specifications require a significantly larger number of operators to describe, whereas the tree structure of TBT permits significantly more concise, easier-to-specify task descriptions.

\begin{figure}[ht]
\centering
\resizebox{\linewidth}{!}{%
\begin{tikzpicture}[
    panel/.style={draw=black!20, rounded corners=5pt, fill=black!1},
    title/.style={font=\small\bfseries, anchor=west},
    tbtnode/.style={draw=black!35, rounded corners=2pt, minimum width=0.82cm, minimum height=0.42cm, font=\scriptsize, align=center, fill=white, inner sep=1.5pt},
    leaf/.style={tbtnode, fill=yellow!18},
    opseq/.style={tbtnode, fill=blue!12, font=\scriptsize\bfseries},
    oppar/.style={tbtnode, fill=green!15, font=\scriptsize\bfseries},
    arr/.style={-{Stealth[length=4pt]}, thick, draw=black!55}
]
\node[panel, minimum width=5.85cm, minimum height=3.7cm, anchor=north west] (tbtpanel) at (0,0) {};
\node[panel, minimum width=7.65cm, minimum height=3.7cm, anchor=north west] (ltlpanel) at (6.15,0) {};

\node[title] at (0.25,-0.28) {TBT};
\node[title] at (6.40,-0.28) {LTL};

\node[opseq] (seq) at (2.95,-0.95) {$\nn{Seq}$};
\node[leaf] (start) at (1.45,-1.85) {$\Diamond s$};
\node[oppar] (par) at (2.95,-1.85) {$\nn{Par}_3$};

\node[leaf] (a) at (0.95,-3.12) {$\Diamond a$};
\node[leaf] (b) at (1.95,-3.12) {$\Diamond b$};
\node[leaf] (c) at (2.95,-3.12) {$\Diamond c$};
\node[leaf] (d) at (3.95,-3.12) {$\Diamond d$};
\node[leaf] (e) at (4.95,-3.12) {$\Diamond e$};

\draw[arr] (seq) -- (start);
\draw[arr] (seq) -- (par);
\draw[arr] (par) -- (a);
\draw[arr] (par) -- (b);
\draw[arr] (par) -- (c);
\draw[arr] (par) -- (d);
\draw[arr] (par) -- (e);

\node[draw=orange!25, fill=white, rounded corners=3pt, inner sep=4pt, anchor=north west] at (6.38,-0.72) {%
\begin{minipage}{7.02cm}
\scriptsize
\setlength{\jot}{2.5pt}
\[
\begin{aligned}
\varphi_{\mathrm{LTL}}=\Diamond\Bigl(s \wedge \LTLnext (&
(\Diamond a\wedge\Diamond b\wedge\Diamond c)
\vee(\Diamond a\wedge\Diamond b\wedge\Diamond d)\\
&\vee(\Diamond a\wedge\Diamond b\wedge\Diamond e)
\vee(\Diamond a\wedge\Diamond c\wedge\Diamond d)\\
&\vee(\Diamond a\wedge\Diamond c\wedge\Diamond e)
\vee(\Diamond a\wedge\Diamond d\wedge\Diamond e)\\
&\vee(\Diamond b\wedge\Diamond c\wedge\Diamond d)
\vee(\Diamond b\wedge\Diamond c\wedge\Diamond e)\\
&\vee(\Diamond b\wedge\Diamond d\wedge\Diamond e)
\vee(\Diamond c\wedge\Diamond d\wedge\Diamond e))\Bigr)
\end{aligned}
\]
\end{minipage}
};
\end{tikzpicture}
}
\caption{Comparison of equivalent TBT and LTL specifications. Both describe the desired behavior of ``\textit{reach S, and afterward at least three of the five possible targets}''. While the LTL formula must explicitly enumerate structures, TBT can express them directly because of its tree structure.}
\label{fig:tbt-vs-ltl}
\end{figure}

\subsection{Temporal Behavior Trees as Reward Functions}
\label{sub:tbt-reward}

As TBTs inherit the robustness semantics of STL, we can use the dense reward signal to guide learning, similar to existing work applying STL to RL~\citep{varnai2020stl}. A positive robustness score indicates that a rollout satisfies a TBT specification task, whereas a negative robustness score indicates a violation of that same specification, with the magnitude yielding a dense margin score. 

Using this, we formulate our reward signal as a robustness-based potential over observed prefixes. For a TBT $\tbt$, we compute the \emph{prefix potential} as $\Phi_t(\tbt) = \rob(\trace[0{:}t], \tbt)$, with $\Phi_t(\tbt) = 0$ when the prefix is too short to admit a valid decomposition.  The dense term for the reward is expressed as the discounted potential difference.
\begin{equation}
r_t^\Delta =
\alpha\Bigl(
\gamma_s\,\Phi_t(\tbt)
-
\Phi_{t-1}(\tbt)
\Bigr),\qquad {t = 1,\ldots,N,}
\label{eq:delta-reward}
\end{equation}
with $\alpha$ being a reward scale coefficient and $\gamma_s\in(0,1]$ a shaping discount. This term is the per-step form of quantitative semantics, rewarding policy changes that improve the robustness score of a TBT specification, thereby avoiding the need to convert to a different object or to use external mechanisms to produce a dense reward signal. 

While the STL robustness semantics are directly inherited for leaves, we have to adapt the reward formulation to account for the semantics of the other TBT nodes. For Leaves with safety properties (such as \textit{``never go through A''}), the TBT semantics must account for the fact that, after a violation of a safety property, no future action can repair it~\citep {baier2008principles}. Because the robustness score defined in \cref{eq:delta-reward} is determined by the previous state, this leads to a situation where no change is possible in the robustness, and the delta-based reward signal disappears~\citep{varnai2020stl}. 

Similarly, the semantics of $\nn{Par}_M$ nodes suffer from a similar issue. As the robustness is computed against the weakest child in the quorum of the $\nn{Par}_M$ node, a step improvement in any child other than the one currently holding the $M$-th largest robustness value is masked, and there is no reward for such improvements~\citep{varnai2020stl}. This leads to potentially beneficial policy changes not being properly rewarded. We combat these issues by adding additional terms adjusting the reward signal for these situations.

To provide a dense reward signal even after a safety violation, we introduce a term 
\begin{equation}
r_t^\text{safety} =
\beta \sum_{\LTLglobally\stlf_g \in \mathcal{G}(\tbt)} \min\bigl(0,\ \rob(\trace[t{:}t], \stlf_g)\bigr).
\label{eq:globally-shaping}
\end{equation}
where $\mathcal{G}(\tbt)$ ranges over all leaves of $\tbt$ encoding safety properties. This term imposes a penalty proportional to the extent of the agent's violation of the safety property. Unlike the delta-based approach,  $r_t^\text{safety}$ accumulates the penalty for repeating violations and pushes the policy back towards safe behavior. This provides a dense signal even when there is no way to recover from violating a safety property in a Leaf.

% To be able to provide an informative dense reward despite the safety leaf saturation, we introduce a term 
% \begin{equation}
% r_t^\text{safety} =
% \beta \sum_{\Box\stlf_g \in \mathcal{G}(\tbt)} \min\bigl(0,\ \rob(\trace[t{:}t], \stlf_g)\bigr).
% \label{eq:globally-shaping}
% \end{equation}
% which imposes a penalty based on the extent to which the agent violates the safety property. Unlike the delta-based approach,  $r_t^\text{safety}$ accumulates the penalty for repeating violations and pushes the policy back towards safe behavior.

For $\nn{Par}_M$ nodes, TBTCL exposes the progress of every child as a second potential term with zero terminal value:
\begin{equation}
r_t^\text{par} = \gamma_s\Psi_t - \Psi_{t-1},
\qquad
\Psi_t = \lambda_\mathrm{par}\sum_{i \in \mathcal{C}(\tbt_o)} \Phi_t(\tbt_i)\ \text{ for } 0 \le t < N,
\qquad \Psi_N = 0,
\label{eq:parallel-shaping}
\end{equation}
where $\mathcal{C}(\tbt_o)$ is the set of immediate children of the parallel objective $\tbt$ and $\lambda_\mathrm{par}$ a shaping scale hyperparameter. While quorum robustness remains the satisfaction score, this formulation ensures that an update that improves any child also improves the rewards, and not only for the one that currently holds the quorum back. 

Combining the prefix potential reward and the two terms, \tbtcl{}~provides a potential-based reward directly from the TBT semantics
\begin{equation}
r_t = r_t^\Delta + r_t^\text{safety} + r_t^\text{par},
\label{eq:total-reward}
\end{equation}

We show several properties of the robustness-based reward, starting by showing that maximizing expected return is equivalent to maximizing the expected terminal robustness net of the safety penalty, $U$, and then relating $U$ to the satisfaction of the specification. This gives a satisfaction-probability bound and, under additional assumptions, almost-sure satisfaction by return maximizers.

For a trace $\trace=(s_0,\ldots,s_N)$ we write $G(\trace)=\sum_{t=1}^{N}\gamma_s^{t-1} r_t$ for the return of Eq.~\ref{eq:total-reward} and $S(\trace)=\sum_{t=1}^{N}\gamma_s^{t-1} r_t^\text{safety}\le 0$ for the accumulated safety penalty, and write $
U(\trace)=\alpha\gamma_s^{N}\,\Phi_N(\tbt)+S(\trace)
$ for the discounted terminal potential net of this penalty.
\begin{lemma}[Return identity]
\label{lem:return}
For every trace, $G(\trace)=U(\trace)-c(s_0)$ with $c(s_0)=\alpha\Phi_0(\tbt)+\Psi_0$ a function of the initial state alone, and $U(\trace)\le 0$ whenever $\trace\not\models\tbt$.
\end{lemma}
\begin{proof}
The delta term telescopes

\begin{equation}
\sum_{t=1}^{N}\gamma_s^{t-1}\alpha\bigl(\gamma_s\Phi_t(\tbt)-\Phi_{t-1}(\tbt)\bigr)
=\alpha\sum_{t=1}^{N}\bigl(\gamma_s^{t}\Phi_t(\tbt)-\gamma_s^{t-1}\Phi_{t-1}(\tbt)\bigr)
=\alpha\bigl(\gamma_s^{N}\Phi_N(\tbt)-\Phi_0(\tbt)\bigr),
\label{eq:telescope}
\end{equation}

the parallel term sums to $\gamma_s^{N}\Psi_N-\Psi_0=-\Psi_0$ since $\Psi_N=0$ and the safety term sums to $S(\trace)$. Both initial potentials depend only on $s_0$, and so does $c(s_0)$ as well. As non-satisfying traces produce negative robustness scores~\citep{fainekos2009robustness}, if $\trace\not\models\tbt$ it follows that $\Phi_N(\tbt)\le 0$ which together with $S\le 0$ leads to $U\le 0$.
\end{proof}

As the constant $c(s_0)$ is outside the policy's control, maximizing the expected return maximizes $\mathbb{E}_\pi[U]$. Since $U\le 0$ on violating traces as per robustness semantics, this expectation yields a bound on the satisfaction probability.

\begin{proposition}[Satisfaction-probability bound]
\label{prop:soundness}
Let $U^*=\sup_\trace U(\trace)$ over the feasible traces of a finite-horizon MDP and assume $0<U^*<\infty$. Every policy $\pi$ satisfies
\[
\Pr_\pi(\trace\models\tbt)\;\ge\;\frac{\mathbb{E}_\pi[U(\trace)]}{U^*}
\;=\;\frac{\mathbb{E}_\pi[G]+\mathbb{E}_{\mu_0}[c(s_0)]}{U^*}.
\]
The bound holds for every $\gamma_s\in(0,1]$, $\alpha>0$ and $\beta,\lambda_\mathrm{par}\ge 0$.
\end{proposition}
\begin{proof}
By Lemma~\ref{lem:return}, $U\le 0$ on violating traces, and $U\le U^*$ on satisfying ones by definition of $U^*$. Splitting $\mathbb{E}_\pi[U]$ by whether $\trace\models\tbt$ therefore gives $\mathbb{E}_\pi[U]\le U^*\Pr_\pi(\trace\models\tbt)$. Dividing by $U^*>0$ proves the bound; the equality is Lemma~\ref{lem:return}.
\end{proof}

\begin{corollary}[Almost-sure satisfaction of return maximizers]
\label{cor:soundness}
Under the assumptions of Proposition~\ref{prop:soundness}, let transitions and termination be deterministic and the action set finite. If $\mu_0$-almost every initial state admits a trajectory with $U>0$, then every policy maximizing $\mathbb{E}_\pi[G]$ satisfies $\tbt$ almost surely.
\end{corollary}
\begin{proof}
Each action sequence determines a trajectory from $s_0$, so the finite horizon and action set ensure that $U^*(s_0)=\max_\trace U(\trace)$ over them is attained. Choosing a maximizing sequence for each observed $s_0$ attains $\mathbb{E}_{\mu_0}[U^*(s_0)]$, the upper bound on $\mathbb{E}_\pi[U]=\mathbb{E}_{\mu_0}\bigl[\mathbb{E}_\pi[U\mid s_0]\bigr]$. Since $\mathbb{E}_\pi[U\mid s_0]\le U^*(s_0)$, every maximizer of $\mathbb{E}_\pi[G]=\mathbb{E}_\pi[U]-\mathbb{E}_{\mu_0}[c(s_0)]$ has $\mathbb{E}_\pi[U\mid s_0]=U^*(s_0)$ for $\mu_0$-almost every $s_0$. Applying Proposition~\ref{prop:soundness} conditional on $s_0$, with $0<U^*(s_0)<\infty$, gives $\Pr_\pi(\trace\models\tbt\mid s_0)\ge 1$. A satisfying trajectory without an instantaneous violation has $S=0$ and $U=\alpha\gamma_s^{N}\rob(\trace,\tbt)>0$.
\end{proof}

In stochastic settings, however, this only gives a loose bound as a maximizer of $\mathbb{E}_\pi[U]$ need not maximize the satisfaction probability, since a policy that satisfies with a large margin most of the time and otherwise leads to a violation, contributing a $U$ close to $0$, can outscore one that satisfies surely with a small margin.

\subsection{Temporal Behavior Trees as Curriculum}\label{sub:tbt-curr}
Owing to their tree structure, we can extract a curriculum directly from a TBT without requiring additional inputs or representations. As a TBT is a tree, every subtree of a TBT is a TBT as well. We use this fact and construct a curriculum which iteratively builds up from the leaf nodes of the TBT toward the full specification. Using this, \tbtcl{} trains a single policy throughout the curriculum, with progression through the curriculum corresponding to which subtree of the full TBT specification provides the robustness-based reward. %~(as described in Section~\ref{sub:tbt-reward}).

\begin{figure}[hbt]
\centering
\begin{tikzpicture}[scale=0.80, every node/.style={transform shape},
    tbtnode/.style={draw, rounded corners=2pt, minimum width=1.2cm, minimum height=0.6cm, font=\small},
    stage/.style={draw, rounded corners=3pt, minimum width=1.6cm, minimum height=0.7cm, font=\small},
    arr/.style={-{Stealth[length=5pt]}, thick},
    >=Stealth
]

% TBT Tree (left)
\node[tbtnode, fill=blue!10] (par) at (0, 0) {$\nn{Par}_2$};
\node[tbtnode, fill=green!10] (seq) at (-1.3, -1.2) {$\nn{Seq}$};
\node[tbtnode, fill=orange!10] (safe) at (1.3, -1.2) {$\Box\,\textit{safe}$};
\node[tbtnode, fill=yellow!15] (red) at (-2.1, -2.4) {$\Diamond\,\textit{red}$};
\node[tbtnode, fill=yellow!15] (blue) at (-0.5, -2.4) {$\Diamond\,\textit{blue}$};

\draw[arr] (par) -- (seq);
\draw[arr] (par) -- (safe);
\draw[arr] (seq) -- (red);
\draw[arr] (seq) -- (blue);

\node[font=\small\bfseries, anchor=south] at (0, 0.55) {TBT Specification};

% Arrow: TBT -> DAG
\draw[-{Stealth[length=7pt]}, very thick, decorate,
      decoration={snake, amplitude=1pt, segment length=8pt}]
      (2.1, -1.0) -- node[above, font=\scriptsize] {Alg.~\ref{alg:dag}} (3.7, -1.0);

% Curriculum DAG (middle)
\node[stage, fill=yellow!20] (s1) at (5.0, 0.0) {\emph{reach\_red}};
\node[stage, fill=yellow!20] (s2) at (5.0, -1.2) {\emph{reach\_blue}};
\node[stage, fill=orange!15] (s3) at (7.4, 0.0) {\emph{avoid\_coll.}};
\node[stage, fill=blue!15] (s4) at (6.2, -2.4) {\emph{joint}};

\draw[arr] (s1) -- (s2);
\draw[arr] (s2) -- (s4);
\draw[arr] (s3) -- (s4);

\node[font=\small\bfseries, anchor=south] at (6.2, 0.55) {Curriculum DAG};

% Arrow: DAG -> Schedule
\draw[-{Stealth[length=7pt]}, very thick, decorate,
      decoration={snake, amplitude=1pt, segment length=8pt}]
      (8.3, -1.0) -- node[above, font=\scriptsize] {Alg.~\ref{alg:scheduler}} (10.0, -1.0);

% Training Schedule (right)expose sub-task transitions via 
\node[stage, fill=yellow!20] (l1) at (11.2, 0.0) {\emph{reach\_red}};
\node[stage, fill=yellow!20] (l2) at (11.2, -0.9) {\emph{reach\_blue}};
\node[stage, fill=orange!15] (l3) at (11.2, -1.8) {\emph{avoid\_coll.}};
\node[stage, fill=blue!15] (l4) at (11.2, -2.7) {\emph{joint}};

\node[font=\scriptsize, anchor=east] at ([xshift=-2pt]l1.west) {1};
\node[font=\scriptsize, anchor=east] at ([xshift=-2pt]l2.west) {2};
\node[font=\scriptsize, anchor=east] at ([xshift=-2pt]l3.west) {3};
\node[font=\scriptsize, anchor=east] at ([xshift=-2pt]l4.west) {4};

\draw[arr] (l1) -- (l2);
\draw[arr] (l2) -- (l3);
\draw[arr] (l3) -- (l4);

\node[font=\small\bfseries, anchor=south] at (11.2, 0.55) {Training Schedule};

\end{tikzpicture}
\caption{Going from a TBT specification to a curriculum DAG via Algorithm~\ref{alg:dag}. Each leaf first becomes a stage, $\nn{Seq}$ induces prerequisite edges and $\nn{Par}_2$ creates a joint stage gated on both children. Finally, topological sort yields the linear training schedule executed by Algorithm~\ref{alg:scheduler}.}
\label{fig:dag-extraction}
\end{figure}
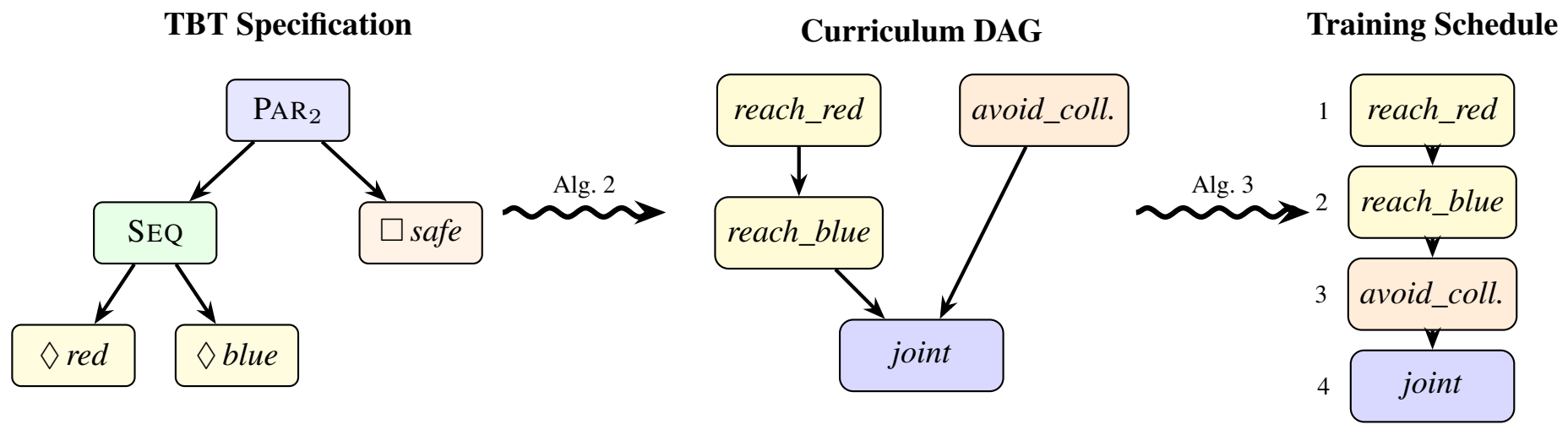

\paragraph{Curriculum Extraction}
The curriculum extraction begins with a walk-through of the TBT, defining the specification task and constructing a Curriculum Graph directly from the semantics of TBTs. We directly use the hierarchical structure of TBTs, traversing through the tree and building a curriculum based on the leaf nodes and the internal nodes connecting them.

First, we compute a Curriculum Graph in which each node encodes a stage (the subtrees specification being learned), and each edge encodes dependencies between stages (which subtask must be mastered before proceeding to the next stage). We convert a TBT into a Curriculum Graph by traversing it, with its semantics governing its encoding.

Each $\nn{Leaf}(\stlf)$ node is converted into a stage, as it corresponds to learning the behavior encoded in the STL formula, such as reaching a target or maintaining distance from a hazard. Sequence nodes  $\nn{Seq}(\tbt_1,\tbt_2)$ impose order between stages and encode a dependency between stages such that $\tbt_1$ must be promoted before the stages extracted from $\tbt_2$ become available. Similarly, $\nn{Fback}(\tbt_1,\ldots,\tbt_k)$ encodes a dependency where a single stage becoming $\nn{Promoted}$ is sufficient to unlock the current stage. 

$\nn{Par}$ and $\nn{Kleene}$ nodes encode \textit{composition} stages as they control how rewards from multiple tasks interact. For $\nn{Par}_M(\tbt_1,\ldots,\tbt_k)$ nodes, every child is first trained as a subobjective, and a joint stage whose objective is the full parallel node becomes available once at least $M$ child stages are promoted. For $\nn{Kleene}(n,\tbt_c)$ nodes, the stages of $\tbt_c$ are followed by one stage per repetition count $k=2,\ldots,n{+}1$, each requiring $k$ repetitions of $\tbt_c$ and unlocked by the promotion of its predecessor, until the full $n{+}1$ repetitions are reached.

We show a pseudo-code implementation in Algorithm~\ref{alg:dag}.

\paragraph{Scheduling}
We convert the Curriculum Graph into a learning schedule by traversing the Curriculum Graph. The scheduler then moves through one stage at a time, tracking the status of each stage in the Curriculum Graph following the progression of  $\nn{Locked} \to \nn{Ready} \to \nn{Active} \to \nn{Promoted}$.

Stages that have no prerequisites, such as those corresponding to $\nn{Leaf}$ nodes, start $\nn{Ready}$ while all others start $\nn{Locked}$. The TBT formula of the stage marked  $\nn{Active}$ is used as the reward function as described in Section~\ref{sub:tbt-reward}.

A stage becomes $\nn{Promoted}$ after the policy has successfully learned the corresponding subtask, meaning that over the sliding window of $W$ episodes either the rolling mean robustness $\overline{\rob}_W$ or the success rate $\eta_W$ is above the respective thresholds ($\overline{\rob}_W > \rob_\text{thresh}\quad\text{or}\quad\eta_W > \eta_\text{thresh}$). We also enforce a minimum number of episodes per stage, $E_\text{min}$, before promotion becomes possible.

When all (or, in the case of $\nn{PAR}_M$, at least $M$) the dependencies of a stage become $\nn{Promoted}$ , it is marked as $\nn{Ready}$ and available for selection by the scheduler. The next stage to become $\nn{Active}$ is selected by a scheduler using an on-the-fly topological sort of the Curriculum Graph.

Once all curriculum stages are promoted, they effectively reconstruct the full TBT specification $\tbt$. \tbtcl{}~ then enters a \emph{consolidation phase} that trains on the full TBT specification, recovering any degraded subtask performance. This phase is further augmented by the \textit{adaptive mode} we present in Section~\ref{sub:tbt-diag}, which actively recovers policies' abilities on specific subtasks by directing the reward towards leaf nodes with the lowest robustness.

We can show that the curriculum that \tbtcl{} builds is a finite, bounded-size DAG (which we discuss in Appendix~\ref{app:proofs}). Furthermore, we can show that the decomposition of the TBT into subtrees, as done during curriculum generation, aligns with the task described in the full TBT specification. While this guarantees neither optimality nor convergence, it shows that the tree structure of TBT can be used to derive a curriculum without requiring external mechanisms or difficulty rankings.

\begin{proposition}[Semantic Curriculum Characterization]
\label{prop:semantic-curriculum}
For any finite TBT specification $\tbt$ and finite trace $\trace$, $\rob(\trace,\tbt)>0$ holds if and only if the operators of $\tbt$ can be resolved recursively into satisfied leaf-level STL obligations on the corresponding trace segments.
\end{proposition}
\begin{proof}
    We use structural induction over the TBT. Leaves directly inherit STL satisfaction and internal nodes follow TBT robustness semantics. Applying the induction hypothesis gives the corresponding satisfied leaf obligations in both directions. The full proof is in Appendix~\ref{app:proofs}.

\end{proof}

\subsection{Temporal Behavior Trees as Diagnostics}\label{sub:tbt-diag}
\begin{figure}[bth]
\centering
\resizebox{\linewidth}{!}{%
\begin{tikzpicture}[
    panel/.style={draw=black!20, rounded corners=5pt, fill=black!1},
    title/.style={font=\small\bfseries, anchor=west},
    tbtnode/.style={draw=black!35, rounded corners=2pt, minimum width=1.15cm, minimum height=0.48cm, font=\scriptsize, align=center, fill=white, inner sep=1.5pt},
    leaf/.style={tbtnode, fill=yellow!18},
    opseq/.style={tbtnode, fill=green!12, font=\scriptsize\bfseries},
    oppar/.style={tbtnode, fill=blue!12, font=\scriptsize\bfseries},
    safeleaf/.style={tbtnode, fill=orange!15},
    arr/.style={-{Stealth[length=5pt]}, thick, draw=black!55},
    bar/.style={draw=black!35, rounded corners=2pt, minimum height=0.28cm, inner sep=0pt, anchor=west},
    lane/.style={font=\scriptsize, anchor=east},
    tick/.style={font=\scriptsize, anchor=north}
]
\node[panel, minimum width=4.65cm, minimum height=3.3cm, anchor=north west] (tbtpanel) at (0,0) {};
\node[panel, minimum width=8.75cm, minimum height=3.3cm, anchor=north west] (tracepanel) at (5.05,0) {};

\node[title] at (0.25,-0.28) {TBT Specification};
\node[title] at (5.30,-0.28) {Trace Decomposition};

\node[oppar] (par) at (2.30,-0.88) {$\nn{Par}_2$};
\node[opseq] (seq) at (1.42,-1.68) {$\nn{Seq}$};
\node[safeleaf] (safe) at (3.18,-1.68) {$\Box\,\textit{safe}$};
\node[leaf] (red) at (0.82,-2.58) {$\Diamond\,\textit{red}$};
\node[leaf] (blue) at (2.02,-2.58) {$\Diamond\,\textit{blue}$};

\draw[arr] (par) -- (seq);
\draw[arr] (par) -- (safe);
\draw[arr] (seq) -- (red);
\draw[arr] (seq) -- (blue);

\draw[-{Stealth[length=6pt]}, very thick, draw=black!50]
      (4.1,-1.95) -- node[above, pos=0.5, font=\scriptsize, align=center, text width=2.6cm]
      {trace segmentation \\mapping\\~\citep{schirmer2024tbt}} (6.14,-1.95);

\coordinate (t0) at (7.45,-0.80);
\coordinate (tk) at (9.72,-0.80);
\coordinate (tT) at (12.62,-0.80);
\draw[-{Stealth[length=4pt]}, draw=black!60] (t0) -- (12.84,-0.80);
\draw[draw=black!45] (t0) -- ++(0,0.10);
\draw[draw=black!45] (tk) -- ++(0,0.10);
\draw[draw=black!45] (tT) -- ++(0,0.10);
\node[tick] at (t0) {$0$};
\node[tick] at (tk) {$k$};
\node[tick] at (tT) {$T$};
\node[font=\scriptsize, anchor=east] at (7.20,-0.80) {$\trace[0{:}T]$};

\draw[densely dashed, draw=black!28] (tk) -- ++(0,-2.4);

\node[lane] at (7.18,-1.25) {$\nn{Par}_2$};
\node[bar, minimum width=5.17cm, fill=blue!18] (parbar) at (7.45,-1.25) {};
\node[font=\scriptsize, anchor=west] at (12.74,-1.25) {$\trace[0{:}T]$};

\node[lane] at (7.18,-1.63) {$\nn{Seq}$};
\node[bar, minimum width=5.17cm, fill=green!16] (seqbar) at (7.45,-1.63) {};
\draw[draw=black!40] (tk |- seqbar.center) -- ++(0,0.15);
\node[font=\scriptsize, anchor=west] at (12.74,-1.63) {$\trace[0{:}T]$};

\node[lane] at (7.18,-2.01) {$\Box\,\textit{safe}$};
\node[bar, minimum width=5.17cm, fill=orange!20] (safebar) at (7.45,-2.01) {};
\node[font=\scriptsize, anchor=west] at (12.74,-2.01) {$\trace[0{:}T]$};

\node[lane] at (7.18,-2.50) {$\Diamond\,\textit{red}$};
\node[bar, minimum width=2.27cm, fill=yellow!30] (redbar) at (7.45,-2.50) {};
\node[font=\scriptsize, anchor=west] at (9.82,-2.50) {$\trace[0{:}k]$};

\node[lane] at (7.18,-2.88) {$\Diamond\,\textit{blue}$};
\node[bar, minimum width=2.90cm, fill=yellow!30] (bluebar) at (9.72,-2.88) {};
\node[font=\scriptsize, anchor=west] at (12.74,-2.88) {$\trace[k{:}T]$};

\end{tikzpicture}
}
\captionsetup{type=figure,hypcap=false}
\caption{Trace decomposition induced by a TBT $\tbt$, mapping subtrees onto subtraces and providing a robustness score for each subtree}
\label{fig:trace-decomposition}
\end{figure}
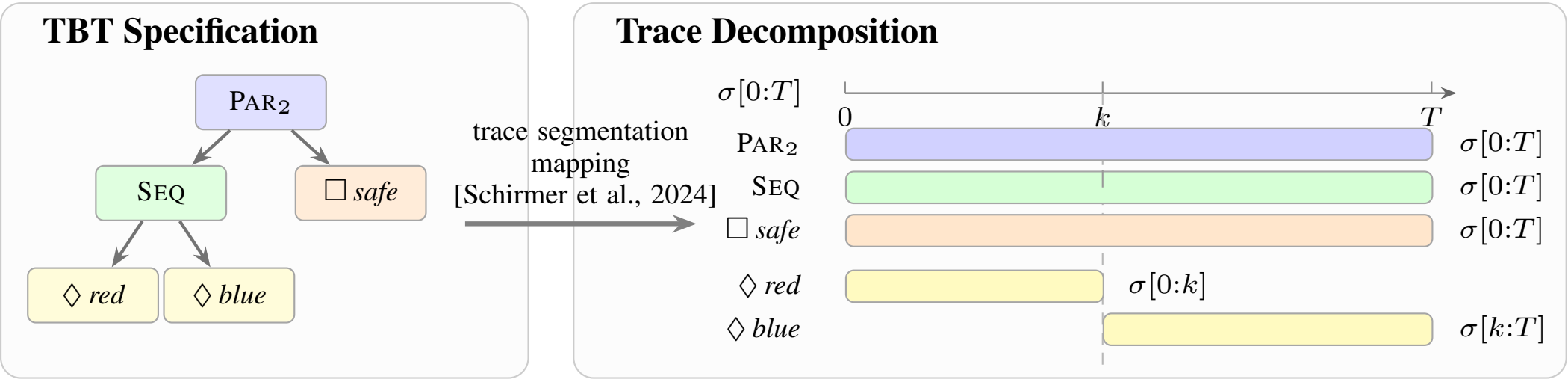

The same semantics of TBTs that make them effective for task specification and learning also enable us to better understand the trained policies. \emph{Trace segmentation}~\citep{schirmer2024tbt} maps every subtree of the TBT to the segment of a rollout on which it was evaluated and assigns that segment a robustness value, as shown in Figure~\ref{fig:trace-decomposition}. While \citet{schirmer2026specifying} use segmentation to repair autonomous drone behavior~\citep{schirmer2026trace}, we use it to examine how a learned policy behaves and how it aligns with the specification described by the TBT.

Aggregating this mapping across several episodes, we can construct a per-leaf competence profile that shows which part of the TBT specification was acted on, when, and how well the policy performed it. This analysis of learned policy behavior follows directly from the formal semantics of the TBTs and their trace segmentation, and provides a more robust analysis of policy actions.

\begin{wrapfigure}{r}{0.5\textwidth}
\vspace{-15pt}% caption baseline level with the first text line
\hfill\begin{minipage}{0.47\textwidth}% 0.03\textwidth extra gutter to the text
\begin{algorithm}[H]
\small\DontPrintSemicolon% reworded for the half-width box (camera-ready)
\caption{Adaptive Training}
\label{alg:adaptive}
\SetKwInOut{Input}{Input}
\Input{trace $\trace$, active stage $s$; window $W_\text{stall}$, budget $R$, threshold $\tau$}
Segment $\trace$ on $s$; record its bottleneck leaf\;
\lIf{$s$ was promoted}{reset stall; end redirect}
\lElseIf{redirect budget $R$ is spent}{resume $s$}
\ElseIf{no promotion for $\geq W_\text{stall}$ episodes}{
  $b^* \gets$ top non-safety bottleneck leaf\;
  \lIf{$\mathrm{freq}(b^*) \geq \tau$}{redirect training to $b^*$}
}
\end{algorithm}
\end{minipage}
\end{wrapfigure}

Furthermore, we can use this technique not only for post-training diagnostics but also to guide training by identifying bottlenecks that occur during learning. By tracking the robustness scores of each leaf over a sliding window of sampled episodes, we can identify \textit{bottleneck nodes} with the lowest robustness scores.

Using this information, during the \textit{consolidation phase}, we can redirect the scheduler to these nodes, using their subspecification to derive the reward and reinforce the policy on this subtask. We call this mechanism \emph{adaptive training}~(Algorithm~\ref{alg:adaptive}), which further mitigates catastrophic forgetting, often observed in compositional learning, by refocusing the agent on specific tasks that degrade during the final phase of learning.

\section{Experiments}
\label{sec:experiments}

\paragraph{Setup}
We evaluate \tbtcl{} on six specifications across two domains, spanning reachability, sequential composition, Kleene repetition, and reach-avoid tasks with safety constraints. We base our implementation on Stable-Baselines3~\citep{raffin2021sb3} and PyTorch~\citep{paszke2019pytorch}, with NumPy~\citep{harris2020array} for the robustness and TBT engines, NetworkX~\citep{hagberg2007exploring} for graph operations, and Hydra~\citep{Yadan2019Hydra} for experiment configuration. We ran the evaluations on a single  Dell R6525 (2x AMD Epyc 7773x - 128 Cores, 256 Threads, 2TB RAM).

\emph{LetterWorld} is a discrete gridworld with lettered target cells; the agent observes its normalized position in $[0,1]^2$ and selects from five actions. Manhattan-distance signals serve as inputs to STL predicates; collisions with obstacles terminate the episode. We use $5 \times 5$ grids for compositional tasks and a $10 \times 10$ grid with 14 obstacles for reach-avoid. \emph{PointNav} is a continuous 2D point-mass domain in $[-5,5]^2$ with continuous acceleration in $[-1,1]^2$; waypoints and circular hazards provide Euclidean distance signals. All methods are trained using PPO~\citep{schulman2017ppo}, $2$M environment steps, 20 random seeds and are evaluated over 100 episodes with the stochastic policy.

We evaluate six specifications (Appendix~\ref{app:specs}): single-goal reachability (\textsc{reach\_red}, \textsc{reach\_a}), sequential composition (\textsc{red\_then\_blue}),  repetition (\textsc{patrol}), and reach-avoid (\textsc{reach\_avoid}). We compare \tbtcl{} against five baselines: \emph{STL Shaping}~\citep{varnai2020stl} (dense robustness reward, no curriculum), \emph{AGCL}~\citep{shukla2023agcl} (DFA-based curriculum, Boolean reward), \emph{SPECTRL}~\citep{jothimurugan2019spectrl} (subpolicy decomposition, sparse reward), \emph{Sparse} ($+1$ on task completion) and  \emph{LDBA}~\citep{hahn2019omega} (using limit-deterministic B\"{u}chi automata to derive the reward).

\begin{figure}[h]
\centering
\input{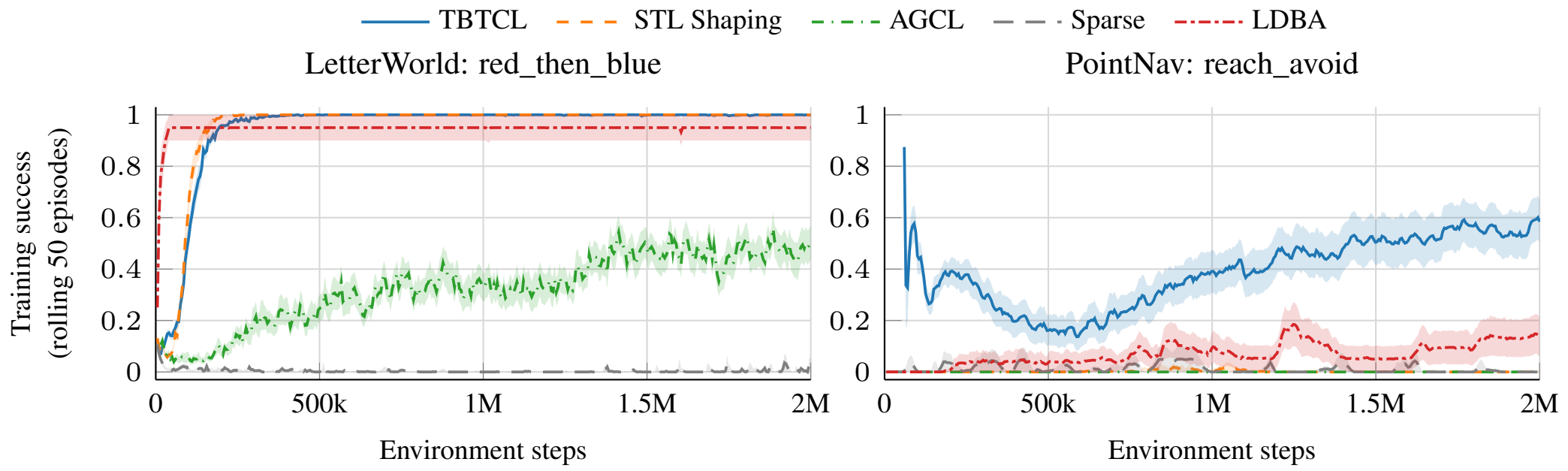}
\caption{Training success on the full specification vs.\ environment steps (rolling mean over 50 episodes, mean $\pm$ SEM over 20 seeds)}
\label{fig:curves}
\end{figure}

\paragraph{Results}
We report the \emph{specification satisfaction rate}: the fraction of episodes where the learned policy produces a trace with positive robustness ($\rob > 0$). A rate of $1.00$ indicates that the policy reliably produces traces that satisfy the full TBT specification. Table~\ref{tab:results} and Figure~\ref{fig:curves} report satisfaction rates on the six specifications we evaluate.

All methods can solve single-goal reachability in LetterWorld. The dense robustness reward drives the sequential and repetition results: \tbtcl{} and STL Shaping both reach $1.00$ on \textsc{red\_then\_blue} and $0.99$ and $1.00$ on \textsc{patrol}, whereas AGCL partially solves both ($0.50$ and $0.60$), LDBA solves the sequence ($0.95$) but not the repetition ($0.40$), and SPECTRL and Sparse fail. The curriculum separates methods on reach-avoid tasks: in LetterWorld, \tbtcl{} achieves $0.85$ versus $0.74$ for STL Shaping and $0.38$ for Sparse; in PointNav, \tbtcl{} reaches $0.64$ while STL Shaping, AGCL, SPECTRL and Sparse remain at $0$ and LDBA reaches $0.14$. (Figure~\ref{fig:curves}).

% generated by scripts/cr/make_table1.py -- do not edit by hand
\begin{table}[h]
\centering
\caption{ Mean specification satisfaction rate (fraction of evaluation episodes with $\rob > 0$) over 20 seeds.}
\label{tab:results}
\small
\setlength{\tabcolsep}{3.5pt}
\begin{tabular}{@{}l cccc cc@{}}
\toprule
 & \multicolumn{4}{c}{\textbf{LetterWorld}} & \multicolumn{2}{c}{\textbf{PointNav}} \\
\cmidrule(lr){2-5} \cmidrule(l){6-7}
\textbf{Method} & \textsc{reach} & \textsc{seq} & \textsc{patrol} & \textsc{avoid} & \textsc{reach} & \textsc{avoid} \\
\midrule
\tbtcl{} & \textbf{1.00}\,$\pm$\,0.00 & \textbf{1.00} & 0.99\,$\pm$\,0.01 & \textbf{0.85\,$\pm$\,0.06} & \textbf{1.00} & \textbf{0.64\,$\pm$\,0.09} \\
\midrule
STL Shaping & \textbf{1.00}\,$\pm$\,0.00 & \textbf{1.00} & \textbf{1.00} & 0.74\,$\pm$\,0.08 & \textbf{1.00} & 0.00\,$\pm$\,0.00$^{}$ \\
AGCL & \textbf{1.00} & 0.50\,$\pm$\,0.07$^{}$ & 0.60\,$\pm$\,0.10$^{}$ & 0.00$^{}$ & 0.00$^{}$ & 0.00$^{}$ \\
SPECTRL & \textbf{1.00} & 0.00\,$\pm$\,0.00$^{}$ & 0.00$^{}$ & 0.38\,$\pm$\,0.11$^{}$ & 0.24\,$\pm$\,0.09$^{}$ & 0.00$^{}$ \\
Sparse & \textbf{1.00} & 0.02\,$\pm$\,0.02$^{}$ & 0.00$^{}$ & 0.38\,$\pm$\,0.11$^{}$ & 0.24\,$\pm$\,0.09$^{}$ & 0.00$^{}$ \\
LDBA  & \textbf{1.00} & 0.95\,$\pm$\,0.05 & 0.40\,$\pm$\,0.11$^{}$ & 0.45\,$\pm$\,0.11 & 0.70\,$\pm$\,0.11$^{}$ & 0.14\,$\pm$\,0.08$^{}$ \\
%LDBA (shaped) & \textbf{1.00} & \textbf{1.00} & \textbf{1.00} & 0.10\,$\pm$\,0.07$^{}$ & 0.40\,$\pm$\,0.11$^{}$ & 0.00$^{}$ \\
\bottomrule
\end{tabular}
\end{table}

These results suggest that dense robustness derived from either STL formula or TBTs, and the curriculum mechanism are complementary. On tasks where the reward provides sufficient information, such as sequential composition and patrol, robustness shaping alone suffices, and the curriculum provides no additional benefit. However, when satisfying multiple subtasks simultaneously requires balancing multiple objectives such as both safety and goal completion, the curriculum becomes essential. It allows the policy to first master reachability subtasks before the safety constraint is introduced, avoiding the reward conflict that stalls all non-curriculum baselines that attempt the full specification at once. The complementary nature of the robustness rewards and the curriculum supports the claim that encoding both objects as a single TBT provides benefit at no additional cost. We provide ablations, failure analyses, and parameter-sensitivity analyses in Appendix~\ref{app:ablation}.

\section{Related Work}
\label{sec:related}
\paragraph{Temporal specifications as rewards and automata.}
STL robustness has been extensively used as a dense reward signal for reinforcement learning and control~\citep{aksaray2016stl,balakrishnan2019stl,varnai2020stl,DBLP:conf/ijcai/CamachoIKVM19}. Automata-based methods, such as Reward Machines~\citep{icarte2018rm,icarte2022rm}, LDBAs~\citep{hahn2019omega} and AGCL~\citep{shukla2023agcl}, expose task structure through automaton states and transitions. We build on this line of work but keep task semantics, reward, curriculum, and diagnostics in a single TBT rather than deriving separate objects or requiring additional inputs. 

\paragraph{Compositional specification languages.}
SPECTRL~\citep{jothimurugan2019spectrl} and DiRL~\citep{jothimurugan2021dirl} also use logical structure to decompose RL tasks, train subpolicies, and compose them at execution time. \tbtcl{} instead trains a single policy through changing subformula rewards, avoiding execution-time policy composition. The TBT semantics also provide trace segmentation, offering direct evidence for policy diagnostics and adaptive consolidation.

\paragraph{Temporal Behavior Trees.}
Temporal Behavior Trees were introduced for robust monitoring and segmentation of cyber-physical traces~\citep{schirmer2024tbt}, and later applied to the analysis of autonomous systems and trace repair~\citep{schirmer2026specifying,schirmer2026trace}. Concurrent work by \citet{puranic2026learning} uses TBTs to shape RL rewards from imperfect demonstrations. Our approach is complementary, as it does not require demonstrations or trace repair; instead, it uses the TBT directly to generate curriculum stages, training rewards, and diagnostics.

\paragraph{Hierarchical and curriculum RL.}
Hierarchical RL decomposes behavior into skills or options~\citep{sutton1999options,dietterich2000maxq,nachum2018hiro,levy2019hac}, while curriculum learning orders training problems to improve optimization~\citep{bengio2009curriculum,narvekar2020curriculum}. \tbtcl{} is closer to curriculum learning: it decomposes the reward and training schedule, not the deployed policy, and obtains the hierarchy from the formal specification rather than from learned option discovery or a task generator~\citep{florensa2017reverse,dennis2020adversarial,jiang2021plr}.

\section{Limitations}

We evaluate \tbtcl{} on six specifications across two small domains, aiming to showcase the potential of this research direction and show that \tbtcl{} matches dense STL shaping on reachability, sequence, and repetition tasks, and improves over baselines on reach-avoid tasks where curriculum and bottleneck redirection matter more. While this is sufficient to show that TBTs can drive rewards, curriculum extraction, and diagnostics within a single RL pipeline, future work should address its scalability, optimization, and adaptation to realistic applications. The current reward construction requires additional terms to handle safety properties and parallel operators, as well as introducing additional hyperparameters. While TBTs provide a dense reward signal, their effective usage requires additional tuning via these parameters. As noted by \citet{schirmer2024tbt}, robustness and trace segmentation can become expensive for long traces and partition-searching operators such as \nn{Seq} and \nn{Kleene}. While our formal results on curriculum structure guarantee a well-formed schedule, they provide no guarantees of convergence or optimal curriculum ordering.

\section{Conclusion}
We presented \tbtcl{}, a framework for using Temporal Behavior Trees as a unified object for specification-guided reinforcement learning. The central idea is that the same TBT should not only define the task but also provide the dense reward, the curriculum structure, and the diagnostic model for learned policies. By adapting TBT robustness to prefix-based rewards, extracting curriculum stages directly from the tree, and using trace segmentation for bottleneck analysis, \tbtcl{} preserves the structure that is lost in existing methods, where specifications must be compiled into separate, dedicated objects for rewards, curricula, and diagnostic tools.  \tbtcl{} matches dense STL shaping on reachability, sequence, and repetition tasks, and improves over baselines on reach-avoid tasks, where curriculum and bottleneck redirection have greater impact. We show the potential of developing a flexible specification language not only for task description but also for derivation of several RL objects at once, providing several benefits at minimal cost.

\begin{ack}
This work was partially supported by the European Union with ERC Grant HYPER (No. 101055412). Views and opinions expressed are however those of the authors only and do not necessarily reflect those of the European Union or the European Research Council Executive Agency.
Neither the European Union nor the granting authority can be held responsible for them.
\end{ack}

\bibliographystyle{unsrtnat}
\bibliography{citations}%%%%%%%%%%%%%%%%%%%%%%%%%%%%%%%%%%%%%%%%%%%%%%%%%%%%%%%%

\appendix
\clearpage

\section{LLM Use Statement}~\label{app:llms}
LLMs (Claude Opus 4.6/4.7/Fable 5) and Grammarly were used for editing and formatting. Agentic Coding tools (Claude Code) were used to implement the codebase and baselines. Other LLM tools (namely Claude Opus 4.6, Gemini 3.1, and GPT-5.5) were used to solicit feedback on the paper draft, proofread the results presented in the paper, and conduct background research on the literature. 
\section{Hyperparameters and Sensitivity}
\label{app:hyperparams}
\begin{center}
\footnotesize
\setlength{\tabcolsep}{4pt}
\begin{tabular}[t]{@{}ll@{}}
\toprule
\textbf{Parameter} & \textbf{Value} \\
\midrule
Algorithm & PPO~\citep{schulman2017ppo} \\
Policy & MLP (SB3 \texttt{MlpPolicy}) \\
Learning rate & $3 \times 10^{-4}$ \\
Rollout length & 2048 steps \\
Minibatch size & 256 \\
Epochs per update & 10 \\
$\gamma$ (PPO discount) & 0.995 \\
Entropy coefficient & 0.01 \\
Max gradient norm & 0.5 \\
Training steps & $2 \times 10^6$ \\
Seeds & 20 (1--20) \\
\bottomrule
\end{tabular}\hfill
\begin{tabular}[t]{@{}ll@{}}
\toprule
\textbf{Curriculum Parameter} & \textbf{Value} \\
\midrule
$\rob_\text{thresh}$ (promotion robustness threshold) & 0.05 \\
$\eta_\text{thresh}$ (promotion success rate threshold) & 0.8 \\
$W$ (sliding window size) & 80 episodes \\
$E_\text{min}$ (minimum episodes before promotion) & 50 \\
$\alpha$ (reward scale) & 5.0 \\
$\beta$ (safety penalty scale) & 0.5 \\
$\lambda_\mathrm{par}$ ($\nn{Par}$ shaping scale) & 0.5 \\
$\gamma_s$ (shaping discount) & 1.0 \\
$\nn{Par}$ term gate & safety-free parallels only \\
Reward clipping & none \\
Kleene cycle window & 100 steps \\
$W_\text{stall}$ (adaptive stall window) & 50 episodes \\
$R$ (redirect budget, episodes) & 30 episodes \\
Redirect frequency threshold $\tau$ & 0.3 \\
\bottomrule
\end{tabular}
\end{center}

\noindent
\begin{minipage}[t]{0.46\textwidth}
\vspace{0pt}
Table~\ref{tab:sensitivity} shows how the performance on two reach-avoid tasks differs in sensitivity to the defaults above, varying one factor at a time (18 settings $\times$ 2 tasks $\times$ 20 seeds $=$ 720 runs). The discrete task is robust to most parameters, with only the reward scale $\alpha$ showing sensitivity ($0.64$ at $\alpha = 1$, $0.98$ at $\alpha = 10$). However, the continuous domain depends on the safety scale ($\beta = 0.1$: $0.20$; $\beta = 1.0$: $0.72$), and on the promotion window ($160$ episodes: $0.42$) the shaping discount ($\gamma_s = 0.995$, matching PPO's discount so that the estimated return is exactly $G$: $0.00$) . %We use $\gamma_s$ as for it the guarantees for the reward hold. 
The effects of the $\nn{Par}_M$ gate and the promotion threshold are limited in our evaluation. Similarly, for the stall window and the redirect budget the impact is overall small but helps with stability and avoiding stalls during learning of complex specifications.
\end{minipage}\hfill
\begin{minipage}[t]{0.52\textwidth}
\vspace{0pt}
\centering
\footnotesize
\setlength{\tabcolsep}{3.5pt}
\captionof{table}{One-factor-at-a-time sensitivity of \tbtcl{} (satisfaction rate, mean $\pm$ SEM over 20 seeds). Bold value = the setting used everywhere else in the paper.}
\label{tab:sensitivity}
% generated by scripts/cr/make_fig_sensitivity.py -- do not edit by hand (needs \usepackage{multirow})
\begin{tabular}{@{}llcc@{}}
\toprule
\textbf{Factor} & \textbf{Value} & LW \textsc{avoid} & PN \textsc{avoid} \\
\midrule
\multirow{4}{*}{$\alpha$ (reward scale)} & 1.0 & 0.64\,$\pm$\,0.08 & 0.52\,$\pm$\,0.10 \\
 & 2.5 & 0.84\,$\pm$\,0.06 & 0.76\,$\pm$\,0.07 \\
 & \textbf{5.0} & 0.85\,$\pm$\,0.06 & 0.64\,$\pm$\,0.09 \\
 & 10.0 & 0.98\,$\pm$\,0.00 & 0.47\,$\pm$\,0.07 \\
\midrule
\multirow{3}{*}{$\beta$ (safety scale)} & 0.1 & 0.85\,$\pm$\,0.06 & 0.20\,$\pm$\,0.07 \\
 & \textbf{0.5} & 0.85\,$\pm$\,0.06 & 0.64\,$\pm$\,0.09 \\
 & 1.0 & 0.85\,$\pm$\,0.06 & 0.72\,$\pm$\,0.08 \\
\midrule
\multirow{3}{*}{PAR scale} & 0.1 & 0.85\,$\pm$\,0.06 & 0.64\,$\pm$\,0.09 \\
 & \textbf{0.5} & 0.85\,$\pm$\,0.06 & 0.64\,$\pm$\,0.09 \\
 & 1.0 & 0.85\,$\pm$\,0.06 & 0.64\,$\pm$\,0.09 \\
\midrule
\multirow{2}{*}{$\gamma_s$ (shaping discount)} & 0.995 & 0.76\,$\pm$\,0.09 & 0.00 \\
 & \textbf{1.0} & 0.85\,$\pm$\,0.06 & 0.64\,$\pm$\,0.09 \\
\midrule
\multirow{2}{*}{PAR gate} & none & 0.89\,$\pm$\,0.05 & 0.70\,$\pm$\,0.09 \\
 & \textbf{safety\_free} & 0.85\,$\pm$\,0.06 & 0.64\,$\pm$\,0.09 \\
\midrule
\multirow{3}{*}{promotion $\rho$ threshold} & 0.01 & 0.91\,$\pm$\,0.04 & 0.62\,$\pm$\,0.10 \\
 & \textbf{0.05} & 0.85\,$\pm$\,0.06 & 0.64\,$\pm$\,0.09 \\
 & 0.2 & 0.90\,$\pm$\,0.04 & 0.65\,$\pm$\,0.09 \\
\midrule
\multirow{3}{*}{promotion window} & 40 & 0.91\,$\pm$\,0.04 & 0.52\,$\pm$\,0.09 \\
 & \textbf{80} & 0.85\,$\pm$\,0.06 & 0.64\,$\pm$\,0.09 \\
 & 160 & 0.85\,$\pm$\,0.06 & 0.42\,$\pm$\,0.09 \\
\midrule
\multirow{3}{*}{stall window} & 25 & 0.83\,$\pm$\,0.07 & 0.68\,$\pm$\,0.09 \\
 & \textbf{50} & 0.85\,$\pm$\,0.06 & 0.64\,$\pm$\,0.09 \\
 & 100 & 0.89\,$\pm$\,0.04 & 0.56\,$\pm$\,0.09 \\
\midrule
\multirow{3}{*}{redirect budget} & 15 & 0.83\,$\pm$\,0.06 & 0.54\,$\pm$\,0.10 \\
 & \textbf{30} & 0.85\,$\pm$\,0.06 & 0.64\,$\pm$\,0.09 \\
 & 60 & 0.93\,$\pm$\,0.04 & 0.59\,$\pm$\,0.10 \\
\bottomrule
\end{tabular}

\end{minipage}

\section{STL and TBT definition}
\label{app:definitions}

\begin{definition}[STL Syntax]
\label{def:stl}
An STL formula is defined inductively as
$\stlf \;{::=}\; \mu \;\mid\; \neg\stlf \;\mid\; \stlf_1 \wedge \stlf_2 \;\mid\; \stlf_1 \vee \stlf_2 \;\mid\; \Diamond_{[l,u]}\stlf \;\mid\; \Box_{[l,u]}\stlf \;\mid\; \stlf_1 \,\mathcal{U}_{[l,u]}\, \stlf_2$,
where $\mu: \mathbb{R}^d \to \mathbb{R}$ is an atomic predicate over the state space and $[l,u]$ are time interval bounds.
\end{definition}

\begin{definition}[STL Robustness]
\label{def:stl-rob}
The quantitative robustness $\rob(\trace, \stlf, t) \in \mathbb{R}$ of a formula $\stlf$ on trace $\trace$ at time $t$ follows the standard recursive definition~\citep{fainekos2009robustness}:
\begin{align}
\rob(\trace, \mu, t) &= \mu(\trace(t)) \label{eq:rob-atomic} \\
\rob(\trace, \neg\stlf, t) &= -\rob(\trace, \stlf, t) \label{eq:rob-neg} \\
\rob(\trace, \stlf_1 \wedge \stlf_2, t) &= \min\bigl(\rob(\trace, \stlf_1, t),\; \rob(\trace, \stlf_2, t)\bigr) \label{eq:rob-and} \\
\rob(\trace, \stlf_1 \vee \stlf_2, t) &= \max\bigl(\rob(\trace, \stlf_1, t),\; \rob(\trace, \stlf_2, t)\bigr) \label{eq:rob-or} \\
\rob(\trace, \Diamond_{[l,u]}\stlf, t) &= \max_{i \in [t+l,\, t+u]} \rob(\trace, \stlf, i) \label{eq:rob-ev} \\
\rob(\trace, \Box_{[l,u]}\stlf, t) &= \min_{i \in [t+l,\, t+u]} \rob(\trace, \stlf, i) \label{eq:rob-glob} \\
\rob(\trace, \stlf_1 \,\mathcal{U}_{[l,u]}\, \stlf_2, t) &= \max_{i \in [t+l,\, t+u]} \min\!\Bigl(\rob(\trace, \stlf_2, i),\; \min_{j \in [t,\, i)} \rob(\trace, \stlf_1, j)\Bigr) \label{eq:rob-until}
\end{align}
When interval bounds $[l,u]$ are omitted, the operators range over the full signal. Maxima and minima over empty index sets are defined as $-\infty$ and $+\infty$ respectively.  Positive robustness of a trace implies Boolean satisfaction and negative robustness implies violation~\citep{fainekos2009robustness}.
\end{definition}

\begin{definition}[TBT Syntax and Robustness]
\label{def:tbt}
Following \citet{schirmer2024tbt} and adding explicit repetition, a TBT node $\tbt$ is defined inductively as
$\tbt \;{::=}\; \nn{Leaf}(\stlf) \;\mid\; \nn{Seq}(\tbt_1, \tbt_2) \;\mid\; \nn{Par}_M(\tbt_1, \ldots, \tbt_k) \;\mid\; \nn{Fback}(\tbt_1, \ldots, \tbt_k) \;\mid\; \nn{Kleene}(n,\tbt_c)$,
where $M$ is the $\nn{Par}$ quorum\rev{,} $\nn{Kleene}(n,\tbt_c)$ requires $n{+}1$ repetitions of $\tbt_c$. Robustness on sub-trace $\trace[l{:}u]$:
\begin{align}
\rob(\trace[l{:}u], \nn{Leaf}(\stlf)) &= \rob(\trace[l{:}u], \stlf) \label{eq:rob-leaf} \\
\rob(\trace[l{:}u], \nn{Seq}(\tbt_1, \tbt_2)) &= \max_{i \in \{l,\ldots,u-1\}} \min\bigl(\rob(\trace[l{:}i], \tbt_1),\; \rob(\trace[i{+}1{:}u], \tbt_2)\bigr) \label{eq:rob-seq} \\
\rob(\trace[l{:}u], \nn{Par}_M(\tbt_1, \ldots, \tbt_k)) &= M\text{-th largest of } \{\rob(\trace[l{:}u], \tbt_j)\}_{j=1}^{k} \label{eq:rob-par} \\
\rob(\trace[l{:}u], \nn{Fback}(\tbt_1, \ldots, \tbt_k)) &= \max_{i \in [l,u]}\; \max_{j \in [1,k]}\; \rob(\trace[{i{:}u}], \tbt_j) \label{eq:rob-fback} \\
\rob(\trace[l{:}u], \nn{Kleene}(n,\tbt_c)) &=
\max_{(\trace_1,\ldots,\trace_{n+1})\in P_{n+1}(l,u)}
\min_{m=1}^{n+1}\rob(\trace_m,\tbt_c)
\label{eq:rob-kleene-app} 
% {\rob(\trace[l{:}u], \star\tbt_c)} &{=
% \max_{k \ge 1}\;\max_{(\trace_1,\ldots,\trace_{k})\in P_{k}(l,u)}
% \min_{m=1}^{k}\rob(\trace_m,\tbt_c)}
% \label{eq:rob-star-app}
\end{align}
$\nn{Seq}$ partitions the trace at an optimal cut point; $\nn{Fback}$ selects the best child on the best suffix. The set $P_{k}(l,u)$ contains the valid ordered partitions of $\trace[l{:}u]$ into $k$ non-empty contiguous segments, and $\max\emptyset = -\infty$ when a segment is too short to be partitioned.
\end{definition}
For a more detailed explanation of TBTs, their semantics, satisfaction relation, and their use in runtime monitoring, we refer the reader to the foundational work by \citet{schirmer2024tbt}.
\section{Curriculum Properties}
\label{app:proofs}
\begin{proof}[Proof of Proposition~\ref{prop:semantic-curriculum} (Semantic Curriculum Characterization)]
We prove the stronger segment-level statement used by structural induction. To avoid side conditions for trace segments that admit no valid cut or partition, we use the standard convention $\max\emptyset=-\infty$. Define $\operatorname{Sat}(\trace[l{:}u],\tbt)$ to mean that the recursive satisfaction conditions induced by $\tbt$ hold on segment $\trace[l{:}u]$:
\begin{align*}
\operatorname{Sat}(\trace[l{:}u],\nn{Leaf}(\stlf))
&\iff
\rob(\trace[l{:}u],\stlf)>0,\\
\operatorname{Sat}(\trace[l{:}u],\nn{Seq}(\tbt_1,\tbt_2))
&\iff
\exists i\in\{l,\ldots,u-1\}:
\operatorname{Sat}(\trace[l{:}i],\tbt_1)\land \operatorname{Sat}(\trace[i{+}1{:}u],\tbt_2),\\
\operatorname{Sat}(\trace[l{:}u],\nn{Par}_M(\tbt_1,\ldots,\tbt_k))
&\iff
\exists Q\subseteq\{1,\ldots,k\},\, |Q|=M:
\forall j\in Q,\; \operatorname{Sat}(\trace[l{:}u],\tbt_j),\\
\operatorname{Sat}(\trace[l{:}u],\nn{Fback}(\tbt_1,\ldots,\tbt_k))
&\iff
\exists j\in\{1,\ldots,k\},\exists i\in\{l,\ldots,u\}:
\operatorname{Sat}(\trace[{i{:}u}],\tbt_j),\\
\operatorname{Sat}(\trace[l{:}u],\nn{Kleene}(n,\tbt_c))
&\iff
\exists(\trace_1,\ldots,\trace_{n+1})\in P_{n+1}(l,u):
\forall m,\; \operatorname{Sat}(\trace_m,\tbt_c).
\end{align*}
We prove by structural induction on $\tbt$ that
\[
\operatorname{Sat}(\trace[l{:}u],\tbt)
\iff
\rob(\trace[l{:}u],\tbt)>0.
\]

\emph{Leaf.}
Equation~\ref{eq:rob-leaf} gives
\[
\rob(\trace[l{:}u],\nn{Leaf}(\stlf))=\rob(\trace[l{:}u],\stlf),
\]
so the equivalence holds by the definition of $\operatorname{Sat}$ for leaves.

\emph{Sequence.}
Assume the claim holds for $\tbt_1$ and $\tbt_2$. By Eq.~\ref{eq:rob-seq},
\[
\rob(\trace[l{:}u],\nn{Seq}(\tbt_1,\tbt_2))
=
\max_{i\in\{l,\ldots,u-1\}}
\min(\rob(\trace[l{:}i],\tbt_1),\rob(\trace[i{+}1{:}u],\tbt_2)).
\]
This value is positive if and only if some cut $i$ makes the displayed minimum positive, which holds iff both child robustness values at that cut are positive. By the induction hypothesis, this is equivalent to $\operatorname{Sat}(\trace[l{:}i],\tbt_1)$ and $\operatorname{Sat}(\trace[i{+}1{:}u],\tbt_2)$ for some valid cut, which is exactly $\operatorname{Sat}$ for the sequence.

\emph{Parallel.}
Assume the claim holds for all children. Equation~\ref{eq:rob-par} says that robustness is the $M$-th largest child robustness value. The $M$-th largest value is positive if and only if at least $M$ children have positive robustness on the shared segment, equivalently, if and only if there exists a size-$M$ quorum of positive children. By the induction hypothesis, this is exactly the existence of a size-$M$ quorum whose children satisfy $\operatorname{Sat}$ on the shared segment.

\emph{Fallback.}
Assume the claim holds for all children. By Eq.~\ref{eq:rob-fback},
\[
\rob(\trace[l{:}u],\nn{Fback}(\tbt_1,\ldots,\tbt_k))
=
\max_{i\in[l,u]}\max_{j\in[1,k]}\rob(\trace[i{:}u],\tbt_j).
\]
This maximum is positive if and only if some child is positive on some suffix. By the induction hypothesis, this is equivalent to $\operatorname{Sat}(\trace[i{:}u],\tbt_j)$ for some child $j$ and suffix start $i$, which is exactly $\operatorname{Sat}$ for fallback. 

\emph{Kleene.}
Assume the claim holds for $\tbt_c$. By Eq.~\ref{eq:rob-kleene-app},
\[
\rob(\trace[l{:}u],\nn{Kleene}(n,\tbt_c))
=
\max_{(\trace_1,\ldots,\trace_{n+1})\in P_{n+1}(l,u)}
\min_{m=1}^{n+1}\rob(\trace_m,\tbt_c).
\]
This value is positive if and only if some valid partition has a positive minimum, which holds if and only if every repetition segment in that partition gives positive robustness for the child. By the induction hypothesis, this is equivalent to every segment in that partition satisfying $\operatorname{Sat}(\trace_m,\tbt_c)$, which is exactly $\operatorname{Sat}$ for $\nn{Kleene}$ nodes.
\end{proof}

\begin{proposition}[Curriculum Well-Formedness]
\label{prop:dag-finite}

For any finite TBT $\tbt$ , Algorithm~\ref{alg:dag} returns a finite acyclic curriculum graph $D=(V_D,E_D)$, with vertex set $V_D$ of at most $|V_D| = |Leaf| + |\nn{Par}| + |\nn{Fback}| + \sum_{\nn{Kleene}(n,\cdot)\in\tbt} n$ nodes and at most  $|E_D| \leq|V_D|^2$ edges.
\end{proposition}
\begin{proof}[Proof of Proposition~\ref{prop:dag-finite} (Curriculum Well-Formedness)]
As \textsc{Extract} visits every node of the finite tree, it leads to a finite curriculum graph. During its traversal, it creates one stage per leaf, one joint stage per $\nn{Par}_M$ node, one gate stage per $\nn{Fback}$ node, none for a $\nn{Seq}$ node, and the $n$ repetition stages $r_2,\ldots,r_{n+1}$ per $\nn{Kleene}(n,\cdot)$ node. By construction, this gives the vertex count.

For acyclicity, we observe that every edge runs from a stage that already exists to one created later. $\nn{Seq}$ connects the latest stage of $\tbt_1$ to the sources of $\tbt_2$, which are created only after $\tbt_1$ has been processed, while $\nn{Par}_M$, $\nn{Fback}$ and $\nn{Kleene}$ connect existing frontiers to the stage created in the same step. By construction, at each step, newly created edges can only point from existing stages to stages created in that step. As they can never encode a dependency in the opposite direction, no directed cycle can exist. The edge bound follows trivially from $E_D\subseteq V_D\times V_D$.
\end{proof}

\section{Ablations and Failure Modes}
\label{app:ablation}

We take a closer look at the components of \tbtcl{} and remove one at a time on the two reach-avoid tasks: the safety term of the reward (Section~\ref{sub:tbt-reward}), the curriculum (Section~\ref{sub:tbt-curr}) and the adaptive training (Section~\ref{sub:tbt-diag}).
The results in Table~\ref{tab:ablation} show that the reward terms and the curriculum are complementary. In PointNav the curriculum without the safety term does not learn at all ($0.00$), and the full reward without a curriculum reaches $0.37$, both significantly below the full method ($0.64$). In LetterWorld the variant without the safety term is identical to the full method over all 20 seeds: a collision ends the episode before the safety leaf is ever violated, so the term never fires and this column measures the curriculum alone, which lifts the satisfaction rate from $0.74$ to $0.85$.

We apply the trace segmentation mechanism from Section~\ref{sub:tbt-diag} to the 100 evaluation episodes of every \tbtcl{} seed on PointNav reach-avoid to analyze how outcomes and bottlenecks vary across seeds. We find that the overall outcome distribution is bimodal: 8 of the 20 seeds satisfy the specification in at least $90\%$ of episodes, 2 never do, and the remaining 10 lie in between. The per-leaf competence profile shows that the safety leaf \emph{avoid\_hazard} is the most common bottleneck for all 12 seeds, and the reach leaf \emph{reach\_a} contributes to issues for 6 seeds. 

% generated by scripts/cr/make_table_ablation.py -- do not edit by hand
\begin{table}[h]
\centering
\caption{Reward-component ablation of \tbtcl{} on the two reach-avoid specifications (satisfaction rate, mean $\pm$ SEM over 20 seeds).}
\label{tab:ablation}
\small
\begin{tabular}{@{}l cc@{}}
\toprule
\textbf{Variant} & LW \textsc{avoid} & PN \textsc{avoid} \\
\midrule
\tbtcl{} (full) & 0.85\,$\pm$\,0.06 & 0.64\,$\pm$\,0.09 \\
\midrule
no safety term ($r^\Delta$ only; PAR term gated off) & 0.85\,$\pm$\,0.06 & 0.00 \\
shaping only (full reward, no curriculum) & 0.74\,$\pm$\,0.08 & 0.37\,$\pm$\,0.10 \\
no adaptive redirect & 0.86\,$\pm$\,0.05 & 0.60\,$\pm$\,0.10 \\
\bottomrule
\end{tabular}
\end{table}

\section{Specification Definitions}
\label{app:specs}

Table~\ref{tab:specs} lists the six specifications evaluated in the main experiments.

\begin{table}[h]
\centering
\caption{Evaluated task specifications. \emph{LW}: LetterWorld (discrete); \emph{PN}: PointNav (continuous).}
\label{tab:specs}
\small
\begin{tabular}{@{}llllr@{}}
\toprule
\textbf{Spec} & \textbf{TBT Formula} & \textbf{Env} & \textbf{Grid / Arena} & $H$   \\
\midrule
\textsc{reach\_red} & $\nn{Leaf}(\Diamond\,(\textit{d\_red} < 1.0))$ & LW & $5 \times 5$ & 50   \\
\textsc{red\_then\_blue} & $\nn{Seq}(\Diamond\,\textit{red},\; \Diamond\,\textit{blue})$ & LW & $5 \times 5$ & 100   \\
\textsc{patrol} & $\nn{Kleene}(2, \nn{Seq}(\Diamond\,\textit{red},\; \Diamond\,\textit{blue}))$ & LW & $5 \times 5$ & 200   \\
\textsc{reach\_avoid} & $\nn{Par}_2(\Diamond\,\textit{red},\; \Box\,(\textit{d\_obs} > 0.5))$ & LW & $10 \times 10$ + obs & 200   \\
\midrule
\textsc{reach\_a} & $\nn{Leaf}(\Diamond\,(\textit{d\_a} < 0.5))$ & PN & $5.0$ & 200 \\
\textsc{reach\_avoid} & $\nn{Par}_2(\Diamond\,\textit{a},\; \Box\,(\textit{d\_hz} > 0.5))$ & PN & $5.0$ + hazards & 300  \\
\bottomrule
\end{tabular}
\end{table}

\section{Algorithm Details}
\label{app:algorithms}

We show the pseudocode implementation of the curriculum extraction and curriculum scheduler algorithms presented in Section~\ref{sub:tbt-curr}.

\begin{algorithm}[h]
\caption{Curriculum DAG Extraction}
\label{alg:dag}
\SetKwInOut{Input}{Input}
\SetKwInOut{Output}{Output}
\SetKwFunction{FExtract}{Extract}
\Input{TBT $\tbt$ }
\Output{Curriculum DAG $D=(V_D,E_D)$, its source stages $A$ and its frontier stages $F$}
\BlankLine
Initialize empty graph $D=(V_D,E_D)$\;
\SetKwProg{Fn}{Function}{:}{}
\Fn{\FExtract{$\tbt$}}{
  \uIf(\tcp*[f]{single leaf stage, Globally leaves included}){$\tbt = \nn{Leaf}(\stlf)$}{
    Create stage $v$ with formula $\stlf$; add $v$ to $V_D$\;
    \Return $(\{v\},\{v\})$\;
  }
  \uElseIf(\tcp*[f]{prefix before suffix}){$\tbt = \nn{Seq}(\tbt_1, \tbt_2)$}{
    $(A_1,F_1) \gets$ \FExtract{$\tbt_1$}; $(A_2,F_2) \gets$ \FExtract{$\tbt_2$}\;
    Add edges $f \to v$ for all $f \in F_1,\, v \in A_2$\tcp*{$\tbt_2$'s sources wait for $\tbt_1$}
    \Return $(A_1, F_2)$\;
  }
  \uElseIf(\tcp*[f]{parts then compose}){$\tbt = \nn{Par}_M(\tbt_1, \ldots, \tbt_k)$}{
    $(A_i,F_i) \gets$ \FExtract{$\tbt_i$} for each child $\tbt_i$\;
    Create joint stage $j$ with formula $\tbt$ and quorum $M$; add $j$ to $V_D$\;
    Add edges $f \to j$ for all $f \in \bigcup_i F_i$\;
    \Return $(\bigcup_i A_i, \{j\})$\;
  }
  \uElseIf(\tcp*[f]{one lap, then $2,\ldots,n{+}1$}){$\tbt = \nn{Kleene}(n, \tbt_c)$}{
    $(A_c,F_c) \gets$ \FExtract{$\tbt_c$}; $P \gets F_c$\;
    \For{$k \gets 2$ \KwTo $n{+}1$}{
      Create repetition stage $r_k$ with formula ``$k$ repetitions of $\tbt_c$'' ($r_{n+1}$ carries $\tbt$ itself); add $r_k$ to $V_D$\;
      Add edges $p \to r_k$ for all $p \in P$; $P \gets \{r_k\}$\;
    }
    \Return $(A_c, P)$\;
  }
  \uElseIf(\tcp*[f]{any child suffices}){$\tbt = \nn{Fback}(\tbt_1, \ldots, \tbt_k)$}{
    $(A_i,F_i) \gets$ \FExtract{$\tbt_i$} for each child\;
    Create gate stage $g$ with formula $\tbt$ and quorum $1$; add $g$ to $V_D$\;
    Add edges $f \to g$ for all $f \in \bigcup_i F_i$\;
    \Return $(\bigcup_i A_i, \{g\})$\;
  }
}
$(A,F) \gets$ \FExtract{$\tbt$}; \Return $(D,A,F)$\;
\end{algorithm}

\begin{algorithm}[h]
\caption{Bottom-Up Curriculum Scheduler}
\label{alg:scheduler}
\SetKwInOut{Input}{Input}
\SetKwProg{Pr}{Procedure}{:}{}
\SetKwFunction{FPromote}{Promote}
\Input{TBT $\tbt$, curriculum DAG $D=(V_D,E_D)$ with stage formulas $X_v$ and quorums $q_v$ , $\rob_\text{thresh}$, $\eta_\text{thresh}$,  $W$, $E_\text{min}$}
Initialize: source stages $\gets$ \nn{Ready}; all others $\gets$ \nn{Locked}\;
\While{training budget not exhausted}{
  $v \gets$ the \nn{Active} stage, else the first \nn{Ready} stage in topological order, set to \nn{Active}\;
  \lIf(\tcp*[f]{consolidation}){no such $v$}{Train one episode on the reward of $\tbt$; \textbf{continue}}
  Train one episode on the reward of $X_v$; record $\rob(\trace, X_v)$ for the episode trace $\trace$\;
  $\overline{\rob}_W \gets$ mean of $\rob$, $\eta_W \gets$ fraction with $\rob > 0$, over the last $W$ episodes recorded for $v$\;
  \lIf{$\mathrm{episodes}(v) \geq E_\text{min}$ \textbf{and} $(\overline{\rob}_W > \rob_\text{thresh}$ \textbf{or} $\eta_W \geq \eta_\text{thresh})$}{\FPromote{$v$}}
}
\Pr{\FPromote{$v$}}{
  $v.\mathrm{status} \gets$ \nn{Promoted}\;
  \ForEach{dependent $d$ of $v$ in $D$ that is \nn{Locked} or \nn{Ready}}{
    \If(\tcp*[f]{quorum met}){at least $q_d$ prerequisites of $d$ are \nn{Promoted}}{
      \lIf{$d$ is a \nn{Fback} gate}{Mark the other prerequisites of $d$ \nn{Skipped}; \FPromote{$d$}}
      \lElse{Mark the other prerequisites of $d$ \nn{Promoted}; $d.\mathrm{status} \gets$ \nn{Ready}}
    }
  }
}
\end{algorithm}

% \newpage
% \clearpage
% \input{checklist.tex}

\end{document}